# FFA Sora, video generation as fundus fluorescein angiography simulator

**Authors:**

Xinyuan Wu[1,6], Lili Wang[2,6], Ruoyu Chen[1], Bowen Liu[1], Weiyi Zhang[1], Xi Yang[3], Yifan Feng[3], Mingguang He[1,4,5**] and Danli Shi[1,4,7*]

**Affiliations:**

1. School of Optometry, The Hong Kong Polytechnic University, Hong Kong.

2. Department of Computing, The Hong Kong Polytechnic University, Hong Kong.

3. Department of Ophthalmology, Zhongshan Hospital, Fudan University, Shanghai, China.

4. Research Centre for SHARP Vision (RCSV), The Hong Kong Polytechnic University, Hong Kong.

5. Centre for Eye and Vision Research (CEVR), 17W Hong Kong Science Park, Hong Kong.

6. These authors contributed equally.

7. Lead contact.

**Correspondence:**

Dr. Danli Shi, The Hong Kong Polytechnic University, Hong Kong, China. Email: danli.shi@polyu.edu.hk (Lead contact)

Prof. Mingguang He, Chair Professor of Experimental Ophthalmology, The Hong Kong Polytechnic University, Hong Kong, China. Email: mingguang.he@polyu.edu.hk

**Summary**

Fundus fluorescein angiography (FFA) is critical for diagnosing retinal vascular diseases, but beginners often struggle with image interpretation. This study develops FFA Sora, a text-to-video model that converts FFA reports into dynamic videos via a Wavelet-Flow Variational Autoencoder (WF-VAE) and a diffusion transformer (DiT). Trained on an anonymized dataset, FFA Sora accurately simulates disease features from the input text, as confirmed by objective metrics: Fréchet Video Distance (FVD) = 329.78, Learned Perceptual Image Patch Similarity (LPIPS) = 0.48, and Visual-question-answering Score (VQAScore) = 0.61. Specific evaluations showed acceptable alignment between the generated videos and textual prompts, with BERTScore of 0.35. Additionally, the model demonstrated strong privacy-preserving performance in retrieval evaluations, achieving an average Recall@K of 0.073. Human assessments indicated satisfactory visual quality, with an average score of 1.570(scale: 1 = best, 5 = worst). This model addresses privacy concerns associated with sharing large-scale FFA data and enhances medical education.

**Keywords:** Fundus fluorescein angiography, Text-to-video generation, Diffusion model, Data sharing, Privacy preservation, Medical education

**Introduction**

Fundus fluorescein angiography (FFA) enables the dynamic visualization of retinal blood flow and lesional changes through the intravenous injection of fluorescein sodium. This technique is critical for assessing blood-retina barrier function and diagnosing various retinal vascular diseases, including diabetic retinopathy (DR) and choroidal neovascularization (CNV).[1-5] Despite its importance, accurately interpreting FFA images demands expertise. For junior ophthalmologists and medical students, the challenge of correlating specific diagnoses with corresponding FFA findings is compounded by the complexity of disease presentations and the limited availability of representative FFA cases.[6]

The growing application of artificial intelligence (AI) offers a potential solution for automatic image interpretation. However, developing and validating these tools requires extensive, high-quality labeled datasets, which can be difficult to obtain. Strict regulations governing the use and exchange of personal health information protect patient privacy but can inadvertently hinder the collaborative data sharing necessary to build effective AI models.[7] Additionally, healthcare providers often work in isolation, making them reluctant to share the clinical data that would improve AI systems. Therefore, innovative strategies are needed to overcome the challenges.

Open AI's Sora has made remarkable progress in generating realistic videos,[8] paving the way for text-to-video generation as a promising solution for synthesizing well-labeled, diversified, and privacy-preserving medical data. Recently, diffusion models (DMs) have gained attention for their capacity to generate high-quality, temporally consistent images and videos through iterative denoising processes.[9] These models have been applied to generate medical data, including chest X-ray images[10], brain MRI scans,[11] multi-modal retinal images,[12] endoscopy videos,[13] and facilitate neurosurgery.[14] These technological advancements have expanded applications across multiple domains. Their potential uses include medical education, clinical training, surgical simulation, medical knowledge dissemination, doctor-patient communication, and biomedical data augmentation.[8] However, generating videos with textual description has yet to be explored for creating lesion-preserving dynamic FFA videos in ophthalmology.

To bridge the gap, we developed FFA Sora, an AI-driven model capable of producing dynamic FFA videos from textual reports. This cross-modality generative model can assist in medical education and offer valuable support for training diagnostic tools and other AI-based models in the future.

**Methods**

We utilized de-identified existing data for our study, which received approval from the Institutional Review Board of the Hong Kong Polytechnic University.

**Dataset**

This study uses a subset of a retrospective FFA dataset.[15] FFA data were obtained from a tertiary hospital in China. The dataset included both the medical records and FFA examinations of patients assessed between November 2016 and December 2019. All FFA videos were captured using Zeiss FF450 Plus and Heidelberg Spectralis systems (Heidelberg, Germany) at a resolution of 768×768 pixels. The collection encompassed a wide variety of ocular conditions, including diabetic retinopathy (DR), retinal vein occlusion (RVO), and central serous chorioretinopathy (CSC). To ensure high-quality data, we implemented a filtering process based on vessel area ratios within the FFA frames, excluding any video whose frame vessel area ratios fell below 0.005.

To address the issue of inconsistent and insufficient frame counts in the original FFA videos, we standardized all videos to a fixed frame count of 21 frames. For videos with more than 21 frames, we selected 21 frames in reverse chronological order. For videos with fewer than 21 frames, additional frames were interpolated to reach the required count. This was achieved using a frame interpolation technique based on linear interpolation, implemented via the OpenCV library.

The process involved extracting all frames from the input video and computing the interpolation positions based on the target frame count. For each interpolated frame, a weighted linear interpolation between two adjacent frames was employed to generate smooth transition frames. Specifically, let an interpolated frame lie between the $i$-th frame and the $(i+1)$-th frame of the original video. The pixel values of the interpolated frame are computed as follows:

$$F_{interpolated} = (1 - t) \times F_i - t \times F_{i+1}$$

where:

$F_i$ and $F_{i+1}$ denote the pixel values of the $i$-th and the $(i+1)$-th frames, respectively, $t$ is the interpolation ratio, which takes values in the range [0,1] and determines the relative contribution of the two adjacent frames to the interpolated frame.

The interpolation ratio $t$ is calculated based on the relative temporal position of the interpolated frame within the sequence. Once all interpolated frames were generated, they were combined with the original frames to produce a standardized video consisting of exactly 21 frames. The result dataset was divided into 80% for training, 10% for validation and 10% for testing.

**Model Architecture**

The framework of FFA Sora was based on the Open-Sora Plan repository ([v1.3.0](v1.3.0)),[16] which we specifically adapted to address the task of generating FFA images from medical textual descriptions. It consisted of two primary components: the Wavelet-Flow Variational Autoencoder (WF-VAE) and the Diffusion Transformer (DiT).

The WF-VAE architecture[17], derived from the Stable-Diffusion Image VAE, was fine-tuned and optimized for the specialized task of text-to-FFA image generation in the medical domain. A key architectural enhancement involved the transformation of Conv2D layers into CausalConv3D layers, enabling the model to process temporal video data, which was essential for the sequential nature of FFA imaging. Furthermore, multi-level wavelet transforms were employed in the encoder to decompose video signals into frequency sub-bands, allowing low-frequency components, which is critical for FFA imaging details, to bypass the backbone network via the Main Energy Flow Pathway, thereby reducing computational redundancy.

The second major component was the DiT architecture,[18] which employed a Transformer-based architecture with cross-modal attention mechanisms to adapt to the requirements of text-to-FFA image generation. The denoising process was performed in a low-dimensional latent space compressed by the VAE, where each Cross-DiT Block, consisting of self-attention, cross-attention, and feed-forward networks, was enhanced with gating mechanisms to refine feature extraction and fusion. This design improved the model's ability to capture both high-level semantics and fine-grained details, ensuring that the generated FFA images accurately reflected clinically relevant features such as global vascular structures and local pathological regions.

**Model Inference**

During the inference phase, the model processed input textual descriptions and generated corresponding FFA images with a resolution of $21 \times 512 \times 512$. The input text typically contained detailed medical descriptions of retinal conditions, such as microaneurysms, non-perfusion area, leakage, and other abnormalities, which were crucial for FFA-based diagnosis. To ensure that the generated images align closely with the described retinal conditions, our model utilized the pre-trained text encoder to capture the semantic nuances of the input effectively. These latent features, conditioned on the text, were passed through the CausalConv3D layers in the decoder, which reconstructed the FFA images by generating temporal and spatial patterns consistent with real FFA images. The inference process employed a tiled convolution approach, which optimized memory usage and ensured high-quality image output. By applying this technique, the model generated coherent and clinically relevant FFA images that accurately depict the specified retinal conditions, while maintaining computational efficiency during inference. The development of FFA Sora is presented in Figure 1.

**Objective Evaluation**

Several objective metrics were employed to assess the quality of the generated FFA videos, including general video evaluation metrics including Fréchet Video Distance (FVD),[19] Learned Perceptual Image Patch Similarity (LPIPS),[20] and Visual-question-answering Score (VQAScore).[21] FVD discerns the likeness in feature distribution between authentic and synthetic videos, encapsulating the holistic excellence and unity. LPIPS, utilizing deep learning-inspired features, measures the perceptual resemblance by comparing image segments, thus revealing nuances beyond the capabilities of conventional pixel-based assessments. VQAScore evaluates the alignment of text-to-video models. Collectively, these evaluation criteria provide a comprehensive analysis of the generated FFA video quality. Together, these metrics provide a comprehensive evaluation of the generated FFA video quality.

For FFA domain-specific evaluation, we used FFA-GPT to translate the generated FFA video into text report and compared it with the original text prompt using Bidirectional Encoder Representations from Transformers Score (BERTScore)[22] as similarity measure. BERTScore utilizes the pre-trained BERT language model to measure conceptual overlap between generated and reference content. Unlike purely lexical metrics, it captures nuanced alignments within broader contexts, providing a deeper assessment of textual congruence.

For privacy preservation evaluation, we aimed to test FFA Sora's ability to generate content-preserving yet deidentifiable data. To achieve this, we employed an image-to-image retrieval method, using Recall@K[23] to measure the proportion of matched visuals retrieved among the top K results. This approach assessed whether videos generated by FFA Sora could be matched with the original patient, ensuring both utility and privacy in the generated content.

**Human Assessment**

This evaluation followed the methodology described previously.[23,24] Three ophthalmologists (X.W., X.Y. and Y.F.) reviewed 50 randomly selected FFA videos generated from the test set, comparing them against the corresponding ground-truth FFA videos. The assessment focused on retinal and vascular structures, lesion integrity, overall coherence, and dynamic range. Each video was rated on a scale from 1 to 5, where 1 represented excellent quality and 5 indicated very poor quality. The criteria were defined as follows: 1 = The modality and lesion characteristics of the generated videos exactly match the text prompts; 2 = The modality of the generated videos corresponds to the text prompts, and the lesion characteristics are basically consistent with the text prompts; 3 = The modality of the generated videos corresponds to the text prompt, and the lesion characteristics are slightly consistent with the text prompts; 4 = The modality of the generated videos corresponds to the text prompt, but the lesion features cannot be generated; 5 = Unable to generate all text-oriented features.

**Results**

The final dataset used in the study consists of 3625 FFA videos paired with 1814 reports, among which 2851 videos with 1429 reports are randomly selected for training, 387 videos with 192 reports, and 387 videos with 193 reports for testing. FFA reports in the dataset include description of various lesions, such as microaneurysms, leakage, neovascularization, capillary non-perfusion, and macular edema (Supplementary Table 1).

Figure 2 demonstrates representative examples of generated FFA videos, including corresponding prompts, ground-truth videos, and the generated video frames. FFA Sora can generate detailed FFA videos from text prompts, illustrating retinal lesions and vascular abnormalities, such as leakage, neovascularization, and microaneurysms, etc. Our model can produce FFA videos for other retinal and choroidal diseases as well, such as uveitis, retinitis pigmentosa (RP), etc. Representative frames of these videos are shown in Supplementary Figure 1.

**Objective evaluation showed excellent quality of generated FFA videos**

The generated videos are comprised of 21 frames that precisely document the entire process of FFA examination, with a particular emphasis on the venous and late phases. The model we employed has exhibited satisfactory performance in producing FFA videos when applied to the test datase, as evidenced by Fréchet Video Distance (FVD) = 329.78, Learned Perceptual Image Patch Similarity (LPIPS) = 0.48, and Visual-question-answering Score (VQAScore) = 0.61 (Table 1A). These metrics serve as pivotal benchmarks in assessing the fidelity and quality of the generated FFA videos.

In addition to the evaluation with above standard generative metrics, we also proposed domain-specific evaluation strategies. Based on previous development of FFA-GPT,[15] a two-stage system which can generate FFA report automatically, our model was evaluated to determine whether the content of the generated videos aligns with the actual characteristics of retinal diseases as denoted by specified textual prompts, with Bidirectional Encoder Representations from Transformers Score (BERTScore) = 0.35 (Table 1A). To provide clarity and visualize these findings, an example of reports by FFA-GPT based on FFA Sora's generated FFA videos are presented in Figure 3.

**Image retrieval demonstrated FFA Sora's excellent performance on privacy preserving**

To further explore FFA Sora's robust feature representation, we investigated its performance in image retrieval tasks, specifically examining the relationship between the generated videos and their ground-truth counterparts. The results of this investigation are summarized in Table 1B, where we report Recall@K scores that include 0.02, 0.04, 0.16 for K = 5, 10, and 50, respectively. Furthermore, the mean recall was calculated to be 0.073, reflecting significant disparities among generated and ground-truth videos. These relatively low Recall@K values indicate that FFA Sora effectively prevents the leakage of confidential image information during video generation.

**Human assessment confirmed high visual quality of generated FFA videos**

A comprehensive analysis was conducted on fifty randomly selected videos produced via our model, which were subjected to meticulous review by three experienced ophthalmologists. This evaluation employed a five-point scale, where a score of 1 denoted excellent quality while a score of 5 indicated very poor quality. The generated videos were systematically compared to the corresponding textual reports and the ground-truth FFA videos, providing a framework for a rigorous assessment of their accuracy and fidelity. Through this evaluative process, the subjective visual quality score averaged at 1.570. The specific results are presented in Table 2, and the average scores and the standard deviations of each ophthalmologist are shown in Supplementary Figure 2.

**Discussion**

In this study, we developed FFA Sora, a text-to-video diffusion transformer designed to generate FFA videos from textual descriptions. The evaluation results demonstrated the model's robustness in generating diverse dynamic abnormalities while retaining localized lesion information. This research showcases the potential of text-to-video models for visualizing FFA reports, offering a novel solution for data sharing and enhancing privacy in machine learning model training and medical AI applications.

Generative models represent important advancements in medical data generation of their ability to produce images of high quality, stability, and diversity.[25-27] Previous research has explored applications of generative adversarial networks (GANs) in retinal image quality enhancement, domain adaptation, retinal vessel segmentation, and cross-modality generation of fundus autofluorescence, FFA, and indocyanine green angioagraphy.[28-34] More recently, diffusion models have been applied to ocular imaging tasks, such as ocular surface structure segmentation for meibomian gland dysfunction grading[35] and data augmentation using retinography images to support deep learning model training.[36] Even though diffusion models beat GANs in image generation,[9] they face limitations of high computational costs, slow sampling processes, and significant resource demands. DiT is, therefore, introduced to improve the performance of diffusion models by replacing the commonly used U-Net backbone with a transformer.[37-39] Building on this foundation, FFA Sora is an advanced application of generative models with a better scalability. This model is the first to apply DiT for creating continuous FFA video content. Comprehensive evaluations, including both objective metrics and expert assessments, confirm that the produced videos preserve diagnostic integrity. This validation underscores the model's potential value within the clinical workflows, contributing to more effective patient management in ocular healthcare.

While diverse datasets are essential for enhancing the accuracy and robustness of AI model training, data sharing is often hindered by rigorous privacy regulations and concerns regarding patient confidentiality, particularly with sensitive medical imaging and clinical videos.[40,41] These challenges limit collaboration and the creation of large, diverse datasets needed for training and validating AI models. Additionally, there is a risk that AI models may inadvertently generate videos by recalling data from their training set, potentially compromising privacy.[42] To address this issue, we conducted an image retrieval evaluation, which revealed notably low Recall@K scores, demonstrating the model's robust privacy-preserving capabilities. By mitigating these privacy concerns, FFA Sora facilitates cross-center data sharing and reproducibility, paving the way for improved collaborative research, clinical training, and the development of AI-driven diagnostic tools.

AI-generated content in the medical domain has a broad implication. Research on AI-assisted medical image generation has demonstrated its value in both medical education[6] and the training of convolutional neural networks (CNNs).[10,43] For instance, Tabuchi H, et al.[6] introduced a text-to-image model to produce ultra-widefield (UWF) retinal color fundus images, significantly improving medical students' diagnostic accuracy. This outcome highlights the potential of integrating cutting-edge AI technologies into contemporary medical education. Compared to their model, FFA Sora offers a more dynamic demonstration of information, thereby facilitating a more comprehensive learning experience for retinal anatomy, vascular perfusion, and pathology for ophthalmologists-in-training. Additionally, FFA Sora's accessibility ensures its broad applicability across educational institutions and healthcare facilities, enabling future ophthalmologists to effectively utilize cutting-edge technologies for diagnosing and managing retinal diseases. As the intersection of AI and medical education continues to evolve, innovations like FFA Sora are poised to play a pivotal role in enhancing the competency of healthcare providers in ophthalmology.

In conclusion, this study presents FFA Sora, a novel modelthat transforms FFA report text into dynamic video representations. These synthetic videos retain critical clinical details while preserving patient privacy, offering a promising solution to persistent data-sharing challenges in healthcare. By enabling secure and efficient visualization, FFA Sora can support collaborative research, enhance clinical training, and accelerate the development of AI-driven diagnostic and medical systems.

**Limitations of the study**

This research has several limitations. First, although FFA Sora demonstrates high accuracy and reliability based on multiple evaluation metrics, its real-world clinical utility remains to be confirmed. Broader external validation using more diverse datasets would help establish its practical value. Second,. although our model demonstrated excellent performance on several metrics such as FVD, our evaluation approach may not fully capture the broader generative quality, thereby requiring more extensive and diversified validation in future work. Finally, comprehensive ophthalmic diagnoses often require multiple imaging modalities, while our current approach is limited to visualizing FFA reports. Integrating multimodal medical imaging could significantly improve AI-based models in ophthalmology, offering more complete diagnostic and treatment support.

**Code availability**

Code is available at https://github.com/PKU-YuanGroup/Open-Sora-Plan/tree/v1.3.0.

**Acknowledgement & Funding**

D.S. and M.H. disclose support for the research and publication of this work from the Global STEM Professorship Scheme (P0046113) and Research Matching Grant Scheme (P0048181) and

**Author contributions**

D.S. conceived the study. L.W. built the text-to-video model. D.S., R.C., B.L., W.Z. conducted the literature search, analyzed the data. L.W. performed objective evaluation. X.W., X.Y., Y.F. completed visual evaluation. X.W., L.W. wrote the manuscript. L.W., X.W. organized figures and tables in this study. M.H. provided the data and facilities. All authors critically revised the manuscript.

**Declaration of interests**

The authors declare no competing interest.

**Supplemental information**

Document S1.

**Supplementary Figure 1: Representative FFA videos of other retinal and choroidal diseases generated by FFA Sora**. RP = retinal pigmentosa, PM = pathological myopia, RAO = retinal artery occlusion.

**Supplementary Figure 2: Specific results of each ophthalmologist. O = Ophthalmologist.** Error bars represent standard deviations of the mean.

**Supplementary Table 1：The main eye conditions extracted from the fundus fluorescein

**angiography reports (total N = 3625).**

**Figure legends**

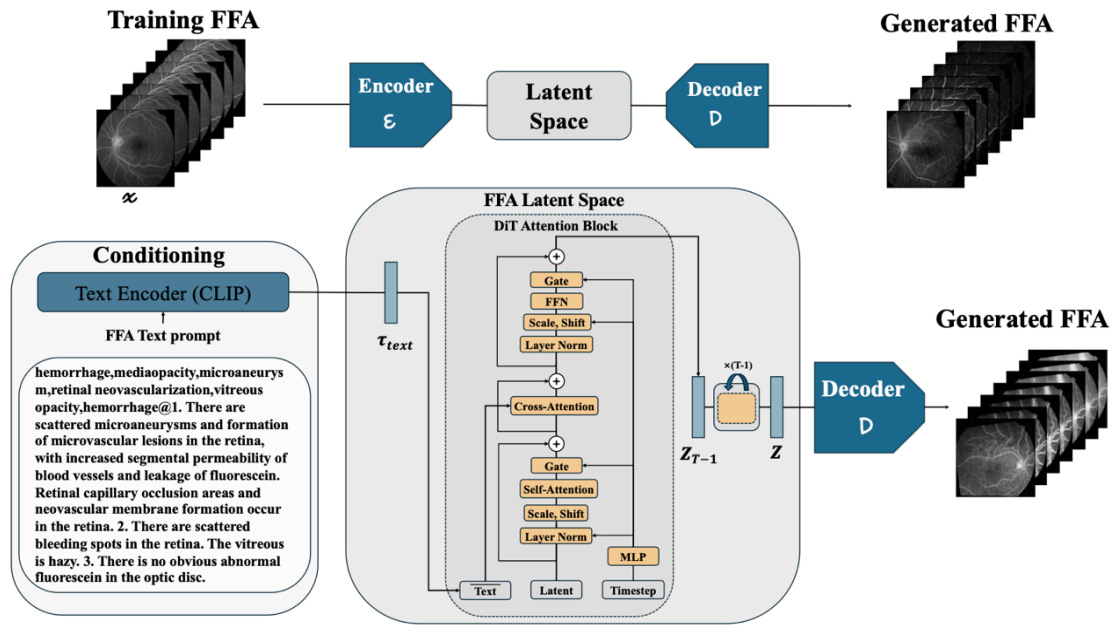

**Figure 1. Development of the text-to-video diffusion model.** This figure illustrates a text-to-video framework for generating Fluorescein Fundus Angiography sequences using a Denoising Diffusion Implicit Transformer. Medical text prompts are encoded by the Text Encoder into a semantic embedding to guide video generation. Input FFA sequences are encoded into latent representations by the Encoder and progressively refined through the DiT Attention Block, which incorporates Self-Attention for spatial/temporal modeling, Cross-Attention for text-to-video conditioning, and timestep embeddings for diffusion-based denoising. The Decoder reconstructs the refined latent representations into realistic FFA video sequences aligned with the input text prompts, enabling clinically meaningful ophthalmic imaging. FFA = fundus fluorescein angiography, DiT = Diffusion Transformer.

**Figure 2. Representative examples of generated FFA videos**. BRVO = branch retinal vein occlusion, CRVO = central retinal vein occlusion, NPDR = non-proliferative diabetic retinopathy, PCV = polypoidal choroidal vasculopathy, AMD = age-related macular degeneration, CSC = central serous chorioretinopathy.

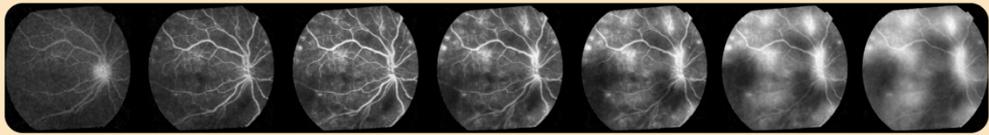

**Figure 3. An example of reports generated by FFA-GPT based on FFA Sora's generated FFA videos.** FFA = fundus fluorescein angiography.

**Tables**

**Table 1. Evaluation results of video quality.**

| A | | | | |
|---|---|---|---|---|
| **Model** | **FVD** | **LPIPS** | **VQAScore** | **BERTScore** |
| FFA Sora | 329.78 | 0.48 | 0.61 | 0.35 |
| **B** | | | | |
| **Image Retrieval** | **Recall@5** | **Recall@10** | **Recall@50** | **Average Recall** |
|  | 0.02 | 0.04 | 0.16 | 0.07 |

A. Video quality evaluation metrics: FVD = Fréchet Video Distance, LPIPS = Learned Perceptual Image Patch Similarity, BERTScore = Bidirectional Encoder Representations from Transformers Score. B. Recall@K and the mean recall. The lower the Recall@K value, the smaller the overlapping part between the generated FFA videos and the ground-truth videos, demonstrating the better performance in preserving patient privacy.

**Table 2. Human assessment results.(n= 50)**

| Human Assessment | O1 | O2 | O3 | Average |
|---|---|---|---|---|
|  | 1.515 | 1.460 | 1.737 | 1.570 |

Human assessments was conducted on a scale of 1 (best) to 5 (worst) for each aspect. O = Ophthalmologist.